\documentclass[sigconf]{acmart}
\usepackage{graphicx} % Required for inserting images
\usepackage{booktabs} % For high-quality tables
\usepackage{xspace}
\usepackage{multirow}
\usepackage{array}
\usepackage{xcolor}
\usepackage{colortbl}
\usepackage{makecell}

\newcommand{\red}[1]{\textcolor{red}{#1}}

\AtBeginDocument{%
  }

\acmConference[ICAIL 2025 - Preprint]{Make sure to enter the correct
  conference title from your rights confirmation email}{June 16--20,
  2025}{Chicago, IL}
\acmSubmissionID{201}

\begin{document}

\title{Automatic Legal Writing Evaluation of LLMs} %: A Brazilian Judicial Case}

\author{Ramon Pires}
\orcid{0000-0002-0023-1971}
\authornotemark[1]
\affiliation{%
  \institution{Maritaca AI}
  \city{Campinas}
  \state{São Paulo}
  \country{Brazil}
}
\email{ramon@maritaca.ai}

\author{Roseval Malaquias Junior}
\orcid{0000-0002-6005-0515}
\affiliation{%
  \institution{Maritaca AI}
  \city{Campinas}
  \state{São Paulo}
  \country{Brazil}
}
\email{roseval@maritaca.ai}

\author{Rodrigo Nogueira}
\orcid{0000-0002-2600-6035}
\affiliation{%
  \institution{Maritaca AI}
  \city{Campinas}
  \state{São Paulo}
  \country{Brazil}
}
\email{rodrigo@maritaca.ai}

% \renewcommand{\shortauthors}{Pires et al.}

\begin{abstract}
    Despite the recent advances in Large Language Models, benchmarks for evaluating legal writing remain scarce due to the inherent complexity of assessing open-ended responses in this domain. 
    % Such evaluation typically requires numerous legal experts and extensive interdisciplinary collaboration to ensure reliable assessment, making it challenging to develop comprehensive benchmarks.
    One of the key challenges in evaluating language models on domain-specific tasks is finding test datasets that are public, frequently updated, and contain comprehensive evaluation guidelines. The Brazilian Bar Examination meets these requirements.
    We introduce \emph{oab-bench}, a benchmark comprising 105 questions across seven areas of law from recent editions of the exam. The benchmark includes comprehensive evaluation guidelines and reference materials used by human examiners to ensure consistent grading.
    We evaluate the performance of four LLMs on \emph{oab-bench}, finding that Claude-3.5 Sonnet achieves the best results with an average score of 7.93 out of 10, passing all 21 exams. 
    We also investigated whether LLMs can serve as reliable automated judges for evaluating legal writing. 
    Our experiments show that frontier models like OpenAI's o1 achieve a strong correlation with human scores when evaluating approved exams, suggesting their potential as reliable automated evaluators despite the inherently subjective nature of legal writing assessment.
    The source code\footnote{\url{https://github.com/maritaca-ai/oab-bench}} and the benchmark\footnote{\url{https://huggingface.co/datasets/maritaca-ai/oab-bench}} -- containing questions, evaluation guidelines, model-generated responses, and their respective automated evaluations -- are publicly available.
\end{abstract}

\begin{CCSXML}
<ccs2012>
   <concept>
       <concept_id>10010147.10010178.10010179.10010182</concept_id>
       <concept_desc>Computing methodologies~Natural language generation</concept_desc>
       <concept_significance>500</concept_significance>
       </concept>
   <concept>
       <concept_id>10010405.10010455.10010458</concept_id>
       <concept_desc>Applied computing~Law</concept_desc>
       <concept_significance>500</concept_significance>
       </concept>
   <concept>
       <concept_id>10002944.10011123.10011130</concept_id>
       <concept_desc>General and reference~Evaluation</concept_desc>
       <concept_significance>300</concept_significance>
       </concept>
 </ccs2012>
\end{CCSXML}

\ccsdesc[500]{Computing methodologies~Natural language generation}
\ccsdesc[500]{Applied computing~Law}
\ccsdesc[300]{General and reference~Evaluation}

\keywords{Open-ended Tasks, Legal Writing, Automatic Evaluation, Brazilian Bar Exam, LLM Judge, Large Language Models}
\begin{teaserfigure}
  \includegraphics[width=\textwidth]{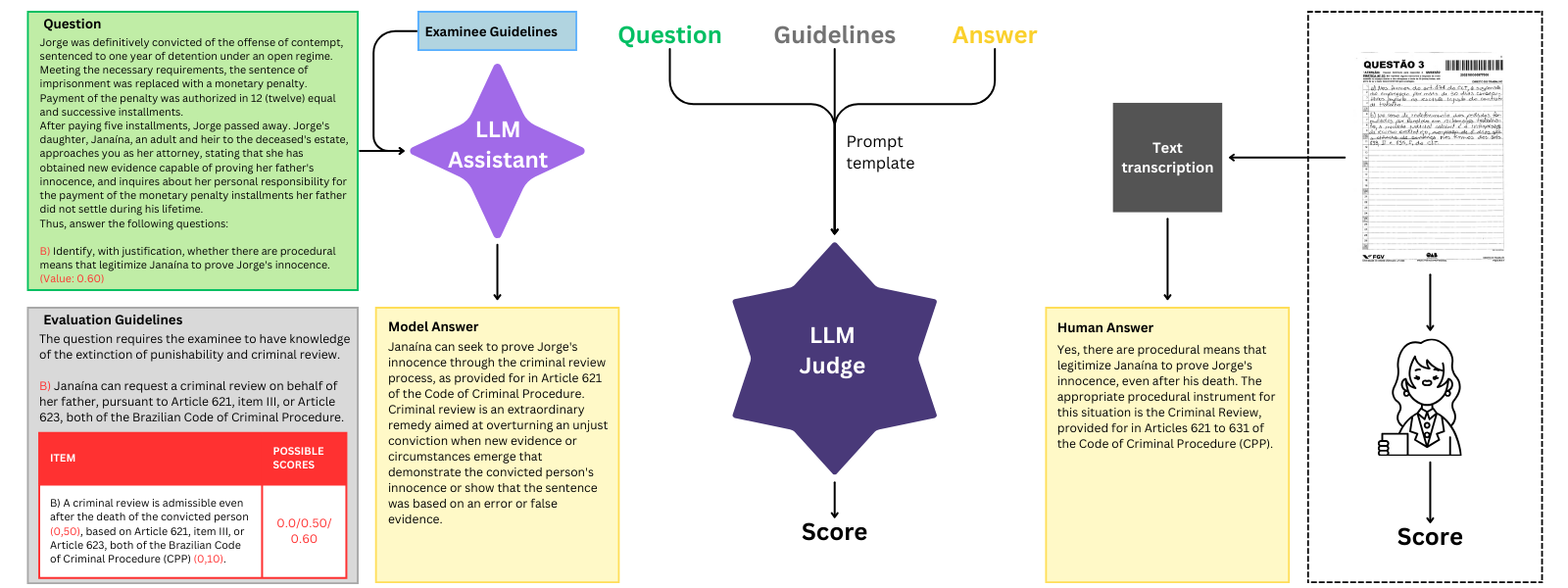}
  \caption{Overview of the \emph{oab-bench} evaluation framework. The left shows the stage of generating responses to exam questions before the assessment. The right shows the collection and conversion of human-written responses from approved candidates to machine-readable format. The middle shows our automated evaluation pipeline that uses LLM judges following official grading guidelines to assess both human and model-generated responses.}
  \Description{Overview of the \emph{oab-bench} evaluation framework.}
  \label{fig:teaser}
\end{teaserfigure}

% \received{28 January 2025}
% \received[revised]{12 March 2025}
% \received[accepted]{5 June 2025}

\maketitle

\section{Introduction}
\label{introduction}
  
Recent advances in large-scale reinforcement learning have enabled Large Language Models (LLMs) to achieve PhD-level performance in standardized exams through reasoning during inference time \cite{OpenAI,deepseekai2025deepseekr1incentivizingreasoningcapability}. However, these emerging capabilities have led to concerns about benchmark saturation, prompting the development of more complex evaluation benchmarks \cite{rein2023gpqagraduatelevelgoogleproofqa,hendrycks2021measuringmassivemultitasklanguage}. % OK.

These efforts mainly focus on STEM benchmarks, where evaluation requires minimal human intervention, typically involving Questions Answering (QA) or short-text generation tasks \cite{jimenez2024swebench, glazer2024frontiermath}. Nonetheless, certain domains remain underexplored for LLM applications due to their inherent evaluation complexity, such as essay writing in specialized domains like the legal domain. In particular, this domain is characterized by the subjective and interpretative nature of some tasks. Evaluating LLM performance in this context often requires a large number of human experts, demanding an expensive interdisciplinary collaboration between machine learning and legal expertise to ensure inter-annotator agreement \cite{fei2023lawbenchbenchmarkinglegalknowledge}. % OK

In this work, we investigate the automatic evaluation of legal open-ended questions using LLMs as judges. We postulate that frontier models can achieve a high correlation with human judges on standardized legal writing exams.
To investigate this, we propose the \emph{oab-bench}, as illustrated in Figure \ref{fig:teaser}. This benchmark comprises 105 questions across seven areas of law, derived from the written portion of the Brazilian Bar Exam. Using the frontier model o1 \cite{OpenAI} as a LLM judge, we generated scores for human responses and compared them with scores provided by human judges for the same questions. Our results demonstrated a strong correlation between the scores assigned by the LLM and those assigned by human judges. % OK

The main contributions of this work are twofold: (1) the \emph{oab-bench}, a benchmark for evaluating legal writing skills of LLMs, addressing the growing need for challenging benchmarks, particularly in the legal domain given the emergence of specialized LLMs \cite{chalkidis-etal-2020-legal,colombo2024saullm7bpioneeringlargelanguage,colombo2024saullm54bsaullm141bscaling}; and (2) an automated evaluation pipeline for this benchmark, supported by our findings that show frontier LLMs, despite the inherently subjective and interpretative nature of the legal writing task, can serve as reliable judges closely aligned with human expert judgment. % OK

\section{Related Work}
\label{related}
In this section, we present studies related to our research proposal. First, we discuss legal benchmarks, focusing on test suites commonly used to evaluate models in closed-ended tasks. We then present studies on the evaluation of open-ended tasks, highlighting their key applications in education and industry with LLM as a judge. % OK

\subsection{Automatic evaluation of LLMs on legal domain} % OK

The rise of domain-specific LLMs trained on legal data \cite{chalkidis-etal-2020-legal,colombo2024saullm7bpioneeringlargelanguage,colombo2024saullm54bsaullm141bscaling} has created a growing need for benchmarks that can automatically evaluate their performance in the legal domain \cite{NEURIPS2023_89e44582,fei2023lawbenchbenchmarkinglegalknowledge,blair-stanek-etal-2024-blt}. Among these, LegalBench \cite{NEURIPS2023_89e44582} stands out for its interdisciplinary collaboration between machine learning experts and legal professionals, ensuring the inclusion of tasks that reflect real-world legal practice. This benchmark comprises 162 tasks, each with at least 50 samples. In particular, seven of these tasks involve open-ended generation, which required evaluation by legal professionals. % OK

However, human evaluation can be inconsistent across different test scenarios, making it labor-intensive to standardize performance assessment. In Chiang et. al~\cite{chiang-lee-2023-large}, the use of LLMs as judges for open-ended story generation and adversarial attacks has been explored. The authors found a correlation between the scores assigned by human evaluators and those given by LLM judges, suggesting that LLMs could replace human evaluation in certain scenarios. % OK

In this context, does this correlation hold for specialized domains such as the legal field? The experiments conducted in this work aim to address this gap. Additionally, to the best of our knowledge, we are the first to provide a dataset of open-ended questions of standardized exams in the Brazilian legal domain. % OK

\subsection{Automatic evaluation of open-ended tasks} % OK

While previous studies have identified a correlation between human evaluators and LLM judges in open-ended task evaluations \cite{chiang-lee-2023-large, liu-etal-2023-g, fu-etal-2024-gptscore}, automatic evaluation still presents limitations that may introduce inconsistencies, such as position bias \cite{wang-etal-2024-large-language-models-fair} and hallucinations \cite{ryu-etal-2023-retrieval}. To address these challenges, recent research has explored techniques to enhance the correlation between human and LLM judges through prompting techniques, such as chain of thought prompting \cite{liu-etal-2023-g} and Retrieval-Augmented Generation (RAG) \cite{ryu-etal-2023-retrieval}. % OK

In particular, Ryu et al. \cite{ryu-etal-2023-retrieval} introduced a framework that uses RAG for the automatic evaluation of open-ended legal QA. Their findings indicated that incorporating relevant legal documents into the evaluation process enhanced the LLM judge alignment with human legal experts. Notably, the LLM judge successfully identified factual errors that were previously overlooked without RAG. This work explores a similar context by evaluating open-ended legal exam questions. However, we hypothesize that alignment with human scores can be further improved by just employing a robust model, such as o1. % OK

Beyond the legal domain, the automatic evaluation of open-ended tasks impacts educational research, where LLMs are used to automatically evaluate essays \cite{MIZUMOTO2023100050,yancey-etal-2023-rating} and provide continuous feedback to learners of a new language \cite{allen2024chatgpt,escalante2023ai}. Additionally, LLMs have been used to evaluate open-ended tasks generated by other LLMs \cite{myrzakhan2024openllmleaderboardmultichoiceopenstylequestions}, similar to this work. The evaluation of LLM performance as chatbots also relies on the assessment of open-ended tasks, as demonstrated in chatbot benchmarks such as AlpacaEval \cite{dubois2024lengthcontrolledalpacaevalsimpleway}. This approach proves particularly effective in generating cost-effective performance signals for reinforcement learning, contributing to the development of post-training datasets such as UltraFeedback \cite{cui2024ultrafeedbackboostinglanguagemodels}. % OK

\section{Methodology}
\label{methodology}

One of the key challenges in evaluating LLMs on domain-specific tasks is finding suitable test datasets that are publicly available, frequently updated to avoid potential training data contamination, and contain comprehensive answer keys and evaluation guidelines to enable automatic evaluations.
The Brazilian Bar Association (OAB) Examination meets these requirements. In this section, we describe our methodology for leveraging these exams and evaluating LLMs as automated judges of legal responses.

\subsection{OAB exam}
\label{exam}

% \todo[inline]{introduce the oab phase 2}
% \todo[inline]{describe the structure of the exams (essay and discursive questions)}
% \todo[inline]{present all the evaluation guidelines (commented answers, score distribution table).}
% \todo{present here a figure of the table, showing an example for which more than one articles are valid}

The OAB examination is a mandatory professional qualification test that all law graduates must pass to practice law in Brazil. The exam is conducted three times a year and consists of two phases: the first phase is an objective test with 80 multiple-choice questions, and the second phase is a dissertative test, requiring candidates to demonstrate their practical legal knowledge through open-ended responses. 
More than 100 thousand candidates take the exam every edition, and 20\% to 45\% are approved.\footnote{\url{https://oab.estrategia.com/portal/estatisticas-completas-do-exame-de-ordem-da-oab/}} 
The OAB examination is organized by the Fundação Getúlio Vargas (FGV). More information on all editions can be accessed through the FGV's official website for the OAB\footnote{\url{https://oab.fgv.br/}}.

In this work, we focus on the second phase of the OAB exam, which comprises a legal essay (peça prático-profissional) worth 5 points and four discursive questions worth 1.25 points each, totaling 10 points. For the legal essay, the candidates must first correctly identify the appropriate type of legal document required based on the case presented, and then write that document, which could be a lawsuit, an appeal, a contract, or any other legal instrument relevant to their chosen law area. The structure of discursive questions comprises two sub-questions labeled A and B.  % Each subquestion has a specific point value assigned, reflecting its weight in the evaluation.

The exam is offered in seven areas of law, with candidates choosing their specialization during registration: Administrative, Civil, Constitutional, Labor, Business, Criminal, and Tax law.
The candidates must achieve a minimum score of 6 points to be approved.

To ensure a standardized evaluation, the examination board provides two key evaluation guidelines: (1) a commented answer with comprehensive explanations of expected legal reasoning and relevant legislation, and (2) a score distribution table that breaks down how scores should be allocated for specific elements in both the essay and the questions. These materials guide examiners in maintaining consistency across evaluations while allowing for flexibility in recognizing alternative valid legal arguments.
Figure~\ref{fig:evaluation_guidelines} presents an example of those evaluation guidelines (translated into English).

\begin{figure}[htb]
    \begin{minipage}{\columnwidth}
        \fontfamily{phv}\selectfont  % Arial font
        % First blue header
        \colorbox{blue!100}{\parbox{\dimexpr\columnwidth-2\fboxsep\relax}{%
            \textcolor{white}{\large\textbf{Commented Answer}}}}
        
        \vspace{0.5em}
        % \normalsize
        % \footnotesize
        \scriptsize
        The question requires the examinee to demonstrate knowledge about \textit{extinction of punishability and criminal review}.
        
        \medskip
        A) No, given the principle of non-transferability of punishment \textbf{\underline{or}} personal responsibility \textbf{\underline{or}} personality \textbf{\underline{or}} non-transmissibility of punishment, the death of the convicted person extinguishes punishability, according to Art. 107, item I, of the Criminal Code, \textbf{\underline{or}} Art. 5, item XLV, of the Federal Constitution \textbf{\underline{or}} Article 5, item 3, of the American Convention on Human Rights - the Pact of San José, Costa Rica (approved by Decree 678/92).
        
        \medskip
        B) Janaína can request a criminal review on behalf of her father, according to Art. 621, item III, \textbf{\underline{or}} Art. 623, both from the Criminal Procedure Code.
        
        \vspace{1em}
        % Second blue header
        \colorbox{blue!100}{\parbox{\dimexpr\columnwidth-2\fboxsep\relax}{%
            \textcolor{white}{\large\textbf{Score Distribution}}}}
            
        \vspace{0.5em}
        % Points table
        \begin{tabular}{|p{0.70\columnwidth}|p{0.22\columnwidth}|}
            \hline
            \rowcolor{red!100}\textcolor{white}{\textbf{ITEM}} & \textcolor{white}{\textbf{SCORE}} \\
            \hline
            A) No, given the principle of non-transferability of punishment \textbf{\underline{or}} personal responsibility \textbf{\underline{or}} personality \textbf{\underline{or}} non-transmissibility of punishment \textcolor{red}{(0.20)}, the death of the convicted person extinguishes punishability \textcolor{red}{(0.35)}, according to Art. 107, item I, of the Criminal Code, \textbf{\underline{or}} Art. 5, item XLV, of the Federal Constitution \textbf{\underline{or}} Article 5, item 3, of the American Convention on Human Rights - the Pact of San José, Costa Rica (approved by Decree 678/92) \textcolor{red}{(0.10)}. & \textcolor{red}{\makecell{0.00/0.20/0.30 \\ 0.35/0.45/0.55 \\ 0.65}} \\
            \hline
            B) Criminal review is admissible even after the death of the convicted person \textcolor{red}{(0.50)}, based on Art. 621, item III, \textbf{\underline{or}} Art. 623, both from the Criminal Procedure Code \textcolor{red}{(0.10)}. & \textcolor{red}{\makecell{0.00/0.50/0.60}} \\
            \hline
        \end{tabular}
    \end{minipage}
    \caption{Example of a commented answer and a score distribution table from the Criminal Law exam (edition 41, question 4). The table shows how the scores are allocated for each item and its parts, with some items accepting multiple valid legal articles as basis.}
    \label{fig:evaluation_guidelines}
\end{figure}

\subsection{The \emph{oab-bench}}
\label{oab_bench}

% \todo[inline]{Describe the release of the benchmark: which editions (39, 40, 41) we use and why}
% \todo[inline]{Present the structure of the dataset}
% \todo[inline]{Share the link (anonymized) to the dataset.}

% The dataset is available at \url{https://huggingface.co/datasets/maritaca-ai/oab-bench}

We introduce \emph{oab-bench}, a benchmark comprising questions and reference materials from three recent editions of the OAB exam (39th, 40th, and 41st), spanning from 2023.3 to 2024.2.
We emphasize that the second phase of these three editions was administered in the year 2024; for instance, the second phase of the 39th edition, which is associated with 2023.3, was actually held at the beginning of 2024.
We chose recent editions to minimize the risk of data contamination in open-source and proprietary LLMs, as these exams were conducted after the known training cutoff dates of the models.

The dataset includes exams from all the seven areas of law covered in the tests. The total number of questions is 105, covering 3 editions $\times$ 7 areas $\times$ 5 questions per exam.
The \emph{oab-bench} includes all the evaluation guidelines needed to judge each question:

\begin{itemize}
  \item The complete question statement
  \item The maximum score possible for that question
  \item Official commented answers and score distribution table
\end{itemize}

The \emph{oab-bench} is available at \url{https://huggingface.co/datasets/maritaca-ai/oab-bench}.

\subsubsection{LLM as a Candidate}
\label{llm_examinee}

The first stage of our evaluation pipeline involves using LLMs to generate answers to the OAB exam questions. 
%To ensure the models approach the task similarly to human candidates, we provide the same instructions given to actual exam takers. 
We provide to the model the same instructions given to actual exam takers. 
Figure~\ref{fig:examinee_prompt} presents the prompt translated into English.

\begin{figure}[htbp]
\begin{center}
% \fbox{
% \begin{minipage}{0.95\columnwidth}\scriptsize\fontfamily{phv}\selectfont
\begin{minipage}{0.95\columnwidth}\colorbox{gray!20}{\parbox{\dimexpr\linewidth-2\fboxsep\relax}{\scriptsize\fontfamily{phv}\selectfont

You are a law graduate taking the second phase of the Brazilian Bar Association (OAB) exam, organized by FGV. Your task is to answer the essay questions and prepare a legal document, demonstrating your legal knowledge, reasoning ability, and skill in applying relevant legislation and jurisprudence to the presented case.

\bigskip

ATTENTION

\medskip

When preparing the texts for the practical-professional document and answers to the essay questions, you must include all necessary data without producing any identification or information beyond what is provided and permitted in the statements contained in the exam booklet. The omission of data that is legally required or necessary for the correct solution of the proposed problem will result in point deductions. You must be careful not to generate any different data that could create an identifying mark.

\medskip

The detection of any identifying mark in the space designated for the transcription of the final texts will result in the annulment of the practical-professional exam and your elimination. For example, when closing the document, you should opt to use only "ellipsis" or "XXX", that is: date "..." or Date "XXX", location "..." or Location "XXX", Attorney "..." or Attorney "XXX", OAB registration "..." or OAB Registration "XXX". Note that in the body of your answers, you should not create any data that generates an identification mark.

\bigskip

OBSERVATIONS

\medskip

PRACTICAL-PROFESSIONAL DOCUMENT: The document must cover all legal grounds that can be used to support the claim. Simply mentioning or transcribing the legal provision does not earn points.

\medskip

QUESTION: You must provide reasoning for your answers. Merely citing the legal provision does not earn points.

\bigskip

From now on, all your answers will compose the final text (not the draft booklet)
}}
\end{minipage}
% }
\caption{Prompt used to instruct the LLM to act as an examinee for the OAB Exam. The prompt includes the same guidelines given to the candidates in the application of the exam.}
\label{fig:examinee_prompt}
\end{center}
\end{figure}

\begin{figure}[htbp]
\begin{center}
% \fbox{
% \begin{minipage}{0.95\columnwidth}\scriptsize\fontfamily{phv}\selectfont
\begin{minipage}{0.95\columnwidth}\colorbox{gray!20}{\parbox{\dimexpr\linewidth-2\fboxsep\relax}{\scriptsize\fontfamily{phv}\selectfont
[Instruction]
\medskip

You are part of the FGV examining board responsible for evaluating law graduates' answers in the second phase of the OAB exam.

The OAB exam grading is analytical. Assign points for each topic covered, considering not only the final result but also legal reasoning, legal basis, and clarity in presenting arguments.

If the candidate presents a partially correct solution but hasn't fully developed the reasoning, it's possible to award a portion of the points for the correct part of the work. This principle aims to value demonstrated legal knowledge, even if the answer isn't perfect.

As a reference to evaluate the answer, use the OAB answer standard, composed of the commented answer and the points distribution table. The commented answer provides examples of valid answers and criteria to be met. The table shows possible scores for each item according to the parts covered. Each item consists of parts, and each part has a score. For example, items representing legal theses are composed of: presentation, argumentation, and legal basis. The SCORE column presents the range of values, from minimum to maximum, for each item.

Begin your evaluation by comparing the candidate's answer with the commented answer. Identify and correct any divergences, striving to be as objective as possible.

Then, analyze each item in the points distribution table, verifying the convergence between the table and the candidate's answer. For each item, check if each part was addressed in the answer, assign 0 or the full score for the respective part (binary evaluation), and provide the accumulated score for the item.

Try not to penalize the candidate when they include articles beyond those required. If the participant references the correct article but their argumentation is incorrect, it's prudent to assign a score of 0.0 for both parts. On the other hand, if the argumentation is coherent and well-constructed but supported by an incorrect article, the argumentation score should be counted.

After analyzing all items, sum up the candidate's total score. Provide the final score, between 0.0 and \textcolor{blue}{\textbf{\{max\_score\}}}, using the following format: "[[score]]". For example, if you assign a score of \textcolor{blue}{\textbf{\{max\_score\}}}, the format would be "[[\textcolor{blue}{\textbf{\{max\_score\}}}]]".

\medskip
[OAB Question]

\textcolor{blue}{\textbf{\{question\}}}

\medskip
[Start of Answer Standard]

\textcolor{orange}{\textbf{\{ref\_answer\_1\}}}

[End of Answer Standard]

\medskip
[Start of Candidate's Answer]

\textcolor{red}{\textbf{\{answer\}}}

[End of Candidate's Answer]
}}
\end{minipage}
% }
\caption{Prompt template used to instruct the LLM to act as an examiner for the OAB Exam. The prompt explains the analytical grading process %, emphasizing the need to evaluate legal reasoning, legal basis, and clarity of arguments. It instructs the model to use the official answer and scoring table as references, 
and stablishes the format of a final score.}
\label{fig:judge_prompt}
\end{center}
\end{figure}

In our experiments we use the temperature of 0.7 to generate model answers.

\subsubsection{LLM as a Judge}
\label{llm_judge}

% \todo[inline]{gives the prompt, describing small details and the sources used to create it.}
% \todo[inline]{describe fastchat and the evaluation pipeline (single, based on scores, multi-turn, etc.)}
% \todo[inline]{describe the judge model o1 and justify the choice}
% \todo[inline]{emphasize that we make availabe the benchmark following the fastchat format, in order to reproduce the experiments (no needed if we not explicitly mention fastchat)}
% \todo[inline]{falar em algum lugar (ainda não sei qual subseção, mas provavelmente antes de apresentar o custo em dólares) que as (1) questões discursivas tem 2 subquestões [done], (2) fazemos avaliação multi-turn quando o item B está sendo avaliado [done], e (2) isso dá um total de 21 * 9 requisições (sendo 9 por exame) [done]}

In this work we use a strong LLM as an examiner to judge model answers~\cite{chiang2023vicuna,zheng2023judging} following the analytical grading process applied in the real world. The judge model receives: (1) the question statement, (2) the maximum possible score for the question, (3) the reference material with the commented answer and the score distribution table, and (4) the answer to be evaluated. The judge then analyzes the candidate's answer against these reference materials to assign appropriate scores based on the demonstrated knowledge and reasoning.

To ensure that the LLM judge behaves similarly to human examiners in the grading process, we investigated online sources (primarily videos) created and shared by legal experts, preparatory course instructors, and candidates approved in previous editions. From these sources, we extracted valuable information to understand how the grading process works in practice, and prepare the judge prompts reflecting those details. % For example, in general the examiners focus on the quality of legal reasoning rather than strict adherence to specific articles -- a well-argued response using an incorrect article reference may receive a partial score, while a poorly reasoned response with correct article citations would not.
Figure~\ref{fig:judge_prompt} presents the resulting prompt template translated into English.

% We use the same platform of MT-Bench, the fastchat llm_judge. 

We use the combination of two judge modes initially proposed by Zheng et al.~\cite{zheng2023judging}: single-answer and reference-guided grading modes. In this scenario, the LLM judge is asked to directly assign a score to a single answer, with access to all the available guidelines or reference answers. Here we use the commented answers and score distribution table to guide the judge (see Section~\ref{exam} for details).

Additionally, to evaluate the second sub-question (labeled as B) of each discursive question, we apply a multi-turn judge prompt.
% By using this approach, we ensure the judge will see the history both the statement and response of the previous subquestion (item A). % this can be relevant because, as the examinee can base his second answer upon information provided in previous sub-question, ... it can be relevant in some aspects.
This is important because sub-questions are often interconnected, where the answer to item B may build upon or reference information provided in item A. By implementing a multi-turn approach, we ensure the judge has the complete context -- both the statement and response from item A -- when evaluating item B. %, allowing for a more accurate assessment that mirrors how human examiners evaluate these connected responses.

The multi-turn mode is complementary to the single-answer mode, as the model still gives a score to a unique answer; and to reference-guided mode, as the judge receives the evaluation guidelines.
% The multi-turn mode complements the other approaches: it maintains the single-answer scoring method while incorporating the reference-guided mode through the evaluation guidelines.

The multi-turn judge prompt is similar to the one illustrated in Figure~\ref{fig:judge_prompt}, except it emphasizes that the model must judge only the candidate's response to the second sub-question, and both the candidate's responses and evaluation guidelines are presented as two separate conversation histories -- one containing the candidate's responses to parts A and B, and another containing the corresponding evaluation guidelines.

We use o1-2024-12-17~\cite{openai2024openaio1card} as the LLM judge, a reasoning model released by OpenAI. O1 was designed to solve complex and critical problems by producing a long chain of thought before responding to the user. The disadvantage is that the model requires more compute to generate several tokens in the thinking stage, reflecting in the cost to the final users.

We experimented with GPT-4o~\cite{openai2024gpt4ocard} as an alternative judge model. However, we observed several limitations that made it unsuitable: it assigned out-of-range scores, failed in computing final scores, and applied inconsistent score scaling that did not match the required intervals.
DeepSeek-R1~\cite{deepseekai2025deepseekr1incentivizingreasoningcapability}, another reasoning model, could be a good alternative as its performance is comparable with o1 and it is open source. However, we observed that it occasionally failed at evaluating the two sub-questions when it should focus only on the last one in multi-turn evaluation.
These issues led us to choose o1 as our judge model.
We show some of those errors in Section~\ref{results} and Appendix~\ref{app:disagreements}.

Evaluating a model on \emph{oab-bench} requires 189 API requests, since we have 21 exams, and each exam needs 9 individual evaluations (1 for the essay and 2 for each discursive question).
The cost with the o1 judge model ranges from \$50 to \$55 USD. % when using Portuguese language.

% We generate model judgments using the temperature of 1.0, as it is the only value supported by the o1 model.
To generate model judgments, for GPT-4o we use temperature 0, while reasoning models (o1 and DeepSeek-R1) require temperature 1.0 as their only supported value.

All the code needed to reproduce the experiments, including the judge prompts and the evaluation pipeline, are available at \url{https://github.com/maritaca-ai/oab-bench}.

\subsection{Comparison of Judges: Human versus LLM}
\label{comparison}

To assess whether the LLM judge effectively performs its role as an examiner, we evaluated real human-written responses using the model. We collected a set of actual answers from law students, already evaluated by human examiners, manually transcribed them from handwritten format, and assessed them independently with the LLM judge. This allowed us to measure the correlation between human and automated scoring, providing insights into the reliability of our automated evaluation approach.

The main challenge in this evaluation was obtaining real exam responses, as there is a scarcity of publicly available graded exams. We addressed this by collecting student responses and their corresponding official scores from materials voluntarily shared on a preparatory course website\footnote{\url{https://www.provadaordem.com.br/blog/post/provas-resolvidas-2a-fase-e-depoimentos-de-aprovados-na-oab/}} by candidates approved in the exam. These materials were available as photographs of handwritten answer booklets, which required careful manual transcription to create machine-readable versions.

While these exams are from older editions, this does not significantly impact our evaluation. The main reason is that the content was shared as images in handwritten format, making it unlikely that the judge model was exposed to this content during training. Even if the model had seen these materials, the risk would be limited to potentially giving higher scores to answers it recognized, but the handwritten image format significantly reduces this possibility.

Despite the relatively small number of exams, we applied two main selection criteria: (1) using at most one exam per area, and (2) seeking responses with different score ranges. Our final selection included:

\begin{itemize}
    \item A Criminal law exam from the 15th edition with perfect scores (10.0/10.0)
    \item A Civil law exam from the 27th edition with an intermediate passing score (6.1/10.0)
    \item A Labor law exam from the 28th edition with a comfortable passing score (8.15/10.0)
\end{itemize}

\section{Results}
\label{results}

\begin{figure}[tbp]
\centering
\includegraphics[width=\columnwidth]{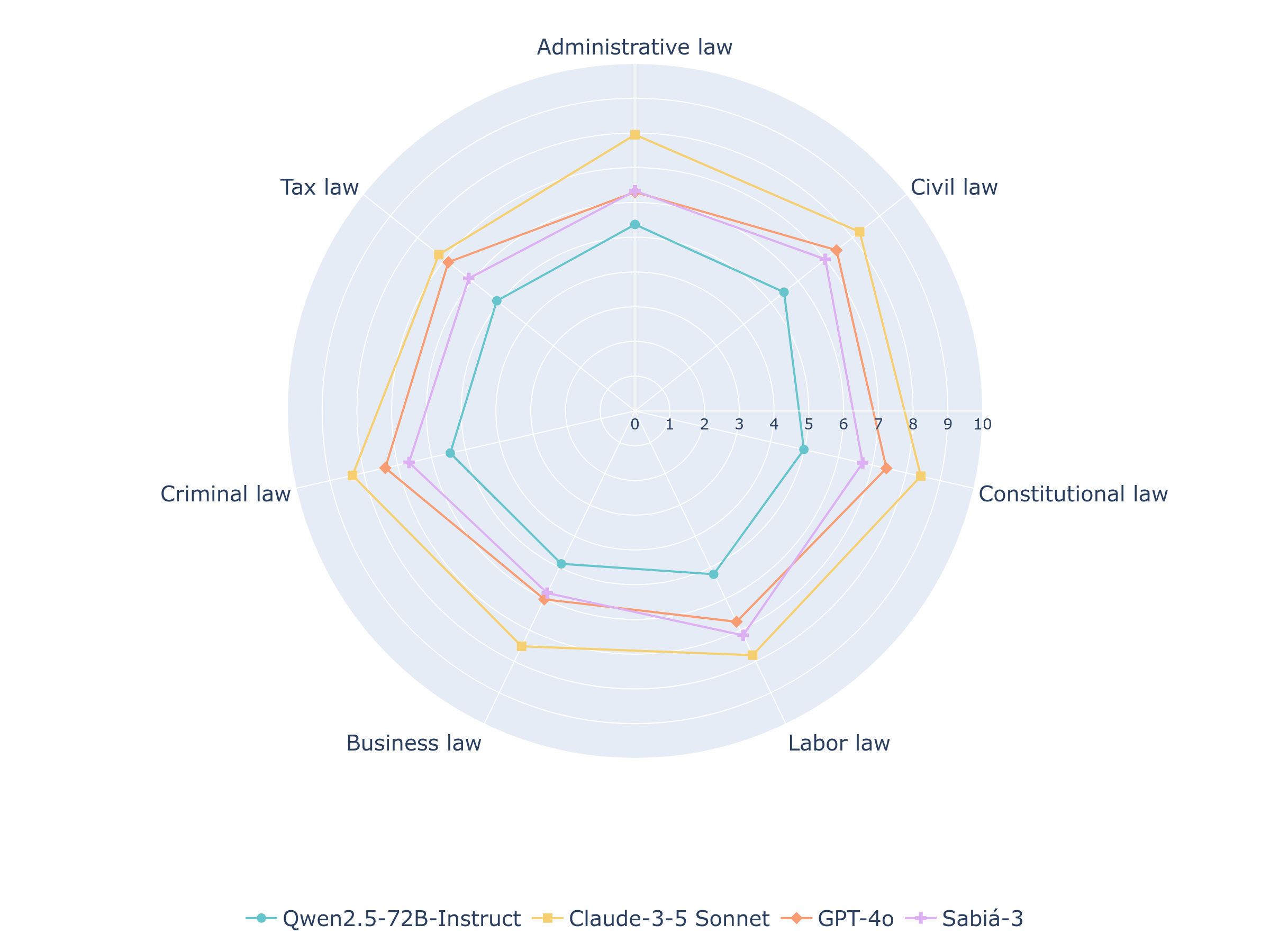}
\caption{Performance of each model across different areas of law.}
\label{fig:polar_plot}
\end{figure}

Table~\ref{tab:results} presents the results of four open-source and proprietary LLMs~\cite{openai2024gpt4ocard,qwen2025qwen25technicalreport,abonizio2024sabia3technicalreport,anthropic2024claude35sonnet} on the \emph{oab-bench}.
For each of the 21 exams, the table shows the total score achieved by each model, with scores ranging from 0 to 10. Scores in red indicate failing grades (below 6.0), which is the minimum passing score required in the actual OAB exam.
Additionally, Figure~\ref{fig:polar_plot} shows the average performance of each model for each law area.

It is important to note that these scores reflect the LLM judge's evaluation and may not align perfectly with how human examiners would grade the same responses. A more comprehensive study involving official examiners would be needed to validate the alignment between LLM and human scoring.

Claude-3-5 Sonnet outperformed other models, achieving an average score of 7.93 across all exams and passing all 21 evaluations. The model excelled particularly in Constitutional law and Criminal law, with average scores of 8.43 and 8.33, respectively. \sloppy

GPT-4o performed second best, with an average score of 6.87 and approval in 18 exams. The model performed particularly well in Civil and Constitutional law, scoring an average of 7.42 in both areas. 
% For Administrative law, the model achieved one of its highest scores (8.30) and also its worst result (3.60), which impacted its overall average.
GPT-4o struggled in Business law, with a marginal score of 6.02.

Sabiá-3 ranked third with an average score of 6.55 (0.32 below GPT-4o), passing 16 of 21 exams. The model showed consistent performance across different law areas, with strong results in Labor law (7.17 average). However, similarly to GPT-4o, Sabiá-3 struggled in Business law, achieving an average score of 5.82 -- the only area below the approval threshold.

% However, similarly to GPT-4o, Sabiá-3 struggled in Business law (5.82), the unique area with the score below the approval threshold. %, as depicted in Figure~\ref{fig:polar_plot}.

Qwen2.5-72B Instruct showed the lowest performance, with an average score of 5.21 and passing only on 5 exams.

\begin{table}[htbp]
\centering
\caption{Results of LLMs on the \emph{oab-bench} benchmark. Scores in red indicate failing grades.}
\label{tab:results}
\resizebox{\columnwidth}{!}{
\begin{tabular}{ll*{4}{>{\centering\arraybackslash}p{\dimexpr0.15\textwidth-2\tabcolsep\relax}}}
% \begin{tabular}{ll>{\centering\arraybackslash}p{2.5cm}>{\centering\arraybackslash}p{2.5cm}>{\centering\arraybackslash}p{2.5cm}>{\centering\arraybackslash}p{2.5cm}}
\toprule
 Edition & Exam & Qwen2.5-72B Instruct & Claude-3-5 Sonnet & GPT-4o & Sabiá-3 \\
\midrule
\multirow{7}{*}{39th}
& Administrative law  & 7.00       & 8.90 & 8.30 & 7.35 \\
& Civil law           & \red{5.30} & 8.35 & 7.00 & 6.20 \\
& Constitutional law  & \red{4.35} & 8.05 & 6.45 & 6.20 \\
& Labor law           & \red{4.50} & 7.45 & 6.10 & 6.55 \\
& Business law        & \red{4.75} & 7.05 & \red{4.65} & 6.20 \\
& Criminal law        & \red{5.75} & 7.95 & 8.15 & 7.80 \\
& Tax law             & 6.55       & 8.40 & 6.95 & 7.15 \\
\midrule
\multirow{7}{*}{40th}
& Administrative law  & \red{4.90} & 8.00 & 7.00 & \red{5.50} \\
& Civil law           & 7.10       & 8.90 & 7.75 & 8.25 \\
& Constitutional law  & \red{4.80} & 9.30 & 8.40 & 7.05 \\
& Labor law           & 6.10       & 8.30 & 7.10 & 8.10 \\
& Business law        & \red{3.95} & 6.90 & 6.00 & \red{5.90} \\
& Criminal law        & \red{3.75} & 7.50 & \red{5.35} & \red{4.95} \\
& Tax law             & \red{3.45} & 6.75 & 6.10 & \red{4.75} \\
\midrule
\multirow{7}{*}{41st}
& Administrative law  & \red{4.20} & 6.95 & \red{3.60} & 6.15 \\
& Civil law           & \red{4.05} & 7.55 & 7.50 & 6.55 \\
& Constitutional law  & \red{5.80} & 7.95 & 7.40 & 6.90 \\
& Labor law           & \red{5.05} & 7.65 & 7.00 & 6.85 \\
& Business law        & \red{5.95} & 8.60 & 7.40 & \red{5.36} \\
& Criminal law        & 6.85       & 9.55 & 8.60 & 7.25 \\
& Tax law             & \red{5.25} & 6.50 & 7.55 & 6.45 \\
\bottomrule
& Mean                & \red{5.21} & 7.93 & 6.87 & 6.55 \\
\bottomrule
\end{tabular}
}
\end{table}

\subsection{Evaluating Human Responses}

This section applies the evaluation pipeline of the \emph{oab-bench} benchmark over real exams collected from the internet (see Section~\ref{comparison}). Here we evaluate the performance with LLM Judges and compare with human examinees.
% we evaluated the performance of human judges by comparing their scores against the official grading guidelines (see Figure~\ref{fig:evaluation_guidelines}).

Table~\ref{tab:judge_comparison} presents the comparison between human and LLM judges across three different exams.
We use Mean Absolute Error (MAE) to measure the average magnitude of differences between human and LLM judge scores. MAE is calculated as $\frac{1}{n}\sum_{i=1}^{n}|y_i - \hat{y_i}|$, where $y_i$ represents the human score and $\hat{y_i}$ represents the LLM judge score for the $i$-th item, providing a straightforward measure of scoring discrepancy.

% From the perspective of LLM judges, both GPT-4o (question 1 of the Criminal law) and DeepSexek-R1 (question 2 of Labor law) demonstrated inconsistencies in scoring, represented with underline in the table. GPT-4o assigned 2.0 points exceeding the maximum possible score of 1.25, while DeepSeek-R1 summed and forced to scale the score until the maximum possible. There discrepancies occurred during the multi-turn evaluation. Despite explicit instructions in the multi-turn prompt to evaluate only the assistant's response for sub-question B within the specified maximum score, the models evaluated from both turns A and B, rather than evaluating only the intended response.

During multi-turn evaluation, two LLM judges showed scoring inconsistencies. In the Criminal law exam (question 1), GPT-4o assigned 2.0 points, exceeding the maximum possible score of 1.25. Similarly, in the Labor law exam (question 2), DeepSeek-R1 had to scale its scores to fit within the maximum limit. These inconsistencies occurred because both models failed to follow the explicit instructions in the multi-turn prompt. Instead of evaluating only the response to sub-question B within its specified maximum score, the models incorrectly evaluated responses from both turns A and B combined.

% In previous experiments we note other mistakes such as forcing a scaling...

Overall, the o1 judge demonstrated reasonable alignment with human scoring, though with some notable variations. O1 model showed slightly better consistency, with MAE ranging from 0.04 to 0.28, while GPT-4o's MAE ranged from 0.20 to 0.55, and DeepSeek-R1's MAE ranged from 0.12 to 0.27.

In the Criminal law exam, the o1 model closely matched the human-graded perfect score of 10.0, while for the Civil law exam, it assigned 7.50, outperforming the human grade of 6.10.
% Finally, the o1 model showed an almost perfect alignment with the human score of 8.15. But by analyzing each question individually, we observe a variation resulting in a MAE of 0.24.
Finally, for the Labor law exam, the o1 model achieved identical total scores, although with slight variations in the distribution across individual questions resulting in a MAE of 0.24.

Considering only the legal essays, the o1 judge tends to assign higher scores than human examiners. %, \todo{improve}{suggesting that it is more generous than human examiners}, except for the maximum-score legal essay penalized in 0.2 by the judge model.
These results suggest that while LLM judges can approximate human scoring patterns, they may require additional constraints or calibration to consistently respect scoring boundaries and maintain human-level precision in borderline cases.

\begin{table}
\centering
\caption{Results Breakdown}
\label{tab:judge_comparison}
\resizebox{\columnwidth}{!}{
\begin{tabular}{cccccccll}
\cmidrule{2-9}
& Judge & Legal essay & Question 1 & Question 2 & Question 3 & Question 4 & Total & MAE \\
                        & & 5.00 & 1.25 & 1.25 & 1.25 & 1.25 & 10.00 & \\
% \cmidrule{2-9}
\midrule
% \rowcolor{gray!20}
Criminal law            & Human         & 5.00 & 1.25 & 1.25 & 1.25 & 1.25 & 10.00 & - \\
% \rowcolor{gray!20}
(15th edition)          & LLM (o1)      & 4.80 & 1.25 & 1.25 & 1.25 & 1.25 & 9.80  {\footnotesize \textcolor{red}{(-0.2)}} & 0.04 \\
% \rowcolor{gray!20}
                        & LLM (GPT-4o)  & 4.60 & \underline{2.00} & 1.25 & 1.25 & 1.25 & \underline{10.35} {\footnotesize \textcolor{blue}{(+0.35)}} & 0.23 \\
% \rowcolor{gray!20}
                        & LLM (DeepSeek-R1)      & 4.40 & 1.25 & 1.25 & 1.25 & 1.25 & 9.40  {\footnotesize \textcolor{red}{(-0.6)}} & 0.12 \\
\midrule
Civil law               & Human         & 3.80 & 0.55 & 0.50 & 0.50 & 0.75 & 6.10 & - \\
(27th edition)          & LLM (o1)      & 4.90 & 0.65 & 0.50 & 0.50 & 0.95 & 7.50  {\footnotesize \textcolor{blue}{(+1.4)}} & 0.28 \\
                        & LLM (GPT-4o)  & 5.00 & 0.00 & 1.05 & 0.75 & 0.95 & 7.75  {\footnotesize \textcolor{blue}{(+1.65)}} & 0.55 \\
                        & LLM (DeepSeek-R1)      & 4.30 & 0.55 & 0.50 & 1.25 & 0.85 & 7.45  {\footnotesize \textcolor{blue}{(+1.35)}} & 0.27 \\
\midrule
% \rowcolor{gray!20}
Labor law               & Human         & 3.90 & 1.25 & 0.60 & 1.15 & 1.25 & 8.15 & - \\
% \rowcolor{gray!20}
(28th edition)          & LLM (o1)      & 4.30 & 1.25 & 0.70 & 1.25 & 0.65 & 8.15  {\footnotesize \textcolor{blue}{(+0)}} & 0.24 \\
% \rowcolor{gray!20}
                        & LLM (GPT-4o)  & 3.50 & 1.25 & 0.70 & 1.15 & 0.75 & 7.35  {\footnotesize \textcolor{red}{(-0.8)}} & 0.20 \\
% \rowcolor{gray!20}
                        & LLM (DeepSeek-R1)      & 4.30 & 1.25 & \underline{0.34} & 1.25 & 0.65 & 7.79 
 {\footnotesize \textcolor{blue}{(+0.36)}} & 0.272 \\
\bottomrule
\end{tabular}
}
\end{table}

Appendix~\ref{app:disagreements} provides a more detailed comparison of judges behavior focused on legal essays.

% \begin{table}
% \centering
% \caption{Results and Mean Absolute Error of LLM-generated answers of 41_direito_civil exam}
% \label{tab:judge_comparison}
% \resizebox{\columnwidth}{!}{
% \begin{tabular}{cccccccll}
% \cmidrule{2-9}
% & Judge & Legal essay & Question 1 & Question 2 & Question 3 & Question 4 & Total & MAE \\
%                         & & 5.00 & 1.25 & 1.25 & 1.25 & 1.25 & 10.00 & \\
% % \cmidrule{2-9}
% \rowcolor{gray!20}
% Qwen2.5-72B Instruct    & Human         & 1.00 & 0.20 & 1.05 & 0.60 & 0.20 & 3.05 ** & - \\
% \rowcolor{gray!20}
%                         & LLM (o1)      & 1.40 & 0.40 & 0.85 & 1.00 & 0.40 & 4.05 {\footnotesize \textcolor{blue}{(+1.0)}} & xxx \\
% % \midrule
% Sabiá-3                 & Human         & 2.00 & 1.05 & 0.25 & 0.60 & 0.80 & 4.70 & - \\
%                         & LLM (o1)      & 3.00 & 1.15 & 0.25 & 1.00 & 1.15 & 6.55 {\footnotesize \textcolor{blue}{(+1.85)}} & xxx \\
% \rowcolor{gray!20}
% % \midrule
% GPT-4o                  & Human         & 2.00 & 0.85 & 0.00 & 1.15 & 0.80 & 4.80 & - \\
% \rowcolor{gray!20}
%                         & LLM (o1)      & 3.35 & 1.15 & 0.85 & 1.25 & 0.90 & 7.50 {\footnotesize \textcolor{blue}{(+2.7)}} & xxx \\
% % \midrule
% % \midrule
% Claude-3-5 Sonnet       & Human         & 2.70 & 0.65 & 1.10 & 1.15 & 0.80 & 6.40 & - \\
%                         & LLM (o1)      & 3.20 & 0.95 & 1.25 & 1.00 & 1.15 & 7.55 {\footnotesize \textcolor{blue}{(+1.15)}} & xxx \\
% \bottomrule
% \end{tabular}
% }
% \end{table}

\section{Limitations}
\label{limitations}

% \todo[inline]{answer generation: not giving to the assistant some tricks that the human candidates know}
% \todo[inline]{cost of judge llm (expensive)}
% \todo[inline]{absence of automatic verification/checking of correct summation, if the score given for a particular item lays down in any possible score, if the model is indeed respecting the variations (different articles allowed), etc.}

Our examinee prompt contains only the standard instructions found in the exam booklet. However, we did not enforce line limits in our model prompts that match the constraints given to human candidates (150 lines for essays and 30 lines per question). Incorporating these constraints could help reduce model verbosity.
% However, human candidates typically go through extensive preparation for the exam, learning specific techniques such as how to structure legal theses with proper formatting -- incorporating facts, legal argumentation, and justification with appropriate articles. While our judge prompt includes specific details about what examiners expect in a well-constructed thesis (see Figure~\ref{fig:judge_prompt}), incorporating these expectations into the LLM assistant prompts could potentially improve their performance.

Additionally, consulting with specialists who are part of the examination body, rather than relying solely on video materials on internet, would enhance the judging criteria with valuable domain expertise. This direct expert input could lead to more precise evaluation guidelines and ensure alignment with current examiners.

With regard to automatic judging, the high cost is a significant obstacle. As of this publication, evaluating a single model on \emph{oab-bench} costs between 50 and 55 USD, which limited our analysis to only four LLMs.
The DeepSeek-R1 could be a promising cost-effective alternative, and with the rapid development of open-source models, we anticipate that future versions may offer even better performance-to-cost ratios. 

Although we measure the correlation of human and LLM judges on real human-written exam responses, the analysis lacks the evaluation of tests that would be reproved.

Furthermore, the current LLM judge pipeline lacks automated validation mechanisms for several critical aspects: there is no automatic verification of score summation accuracy, no system to check if individual item scores fall within their permitted ranges, and no validation to ensure the model properly accounts for allowed variations in article combinations. Implementing such verification systems would add an important layer of quality control to the evaluation process.
Our experiments revealed that GPT-4o occasionally failed in some of these aspects. While such errors were not observed with the o1 judge, we acknowledge that our manual validation was not exhaustive enough to guarantee the complete absence of errors.

\section{Conclusion}
\label{conclusion}

% \todo[inline]{A more extensive study would be required to validate the judge model as an examiner, or a copilot to the human examiners.}
% \todo[inline]{Rapid advance and recent interest on reasoning models could enable more comprehensive evaluations across a broader range of models while maintaining reasonable budget constraints.}

In this work, we introduced \emph{oab-bench}, a benchmark comprising 105 questions across seven areas of law from recent editions of the Brazilian Bar Exam for evaluating the capacity of LLMs in legal writing tasks. Our experiments demonstrated that strong LLMs can achieve considerable performance in this domain, with Claude-3-5 Sonnet passing all 21 exams with an average score of 7.93.
% benchmark for evaluating legal writing skills of LLMs

% We also investigated the feasibility of using LLMs as automated judges for legal writing evaluation. Our results showed that the o1 model achieved reasonable alignment with human scoring, demonstrating consistent evaluation capabilities across human-written and approved tests on three different areas of law.

We also investigated the feasibility of using LLMs as automated judges for evaluating open-ended tasks on legal domain. Our results showed that the o1 model achieved reasonable alignment with human scoring when evaluating approved exams across different areas of law. 
This finding is particularly relevant, as subjective tasks in this domain require a large number of human judges to achieve high inter-annotator agreement.

%% This finding is particularly relevant given that there is a need for a high number of human evaluator in these subjetive tasks to acrquire a high inter-annotator agreement.

% However, a more comprehensive analysis would require testing with failing-grade responses, which were not available in our collected dataset, to have more robust comparisons between human and LLM judges across a broader range of performance levels.
% Additionally, having official examiners evaluate model-generated responses would provide a more accurate assessment of LLM performance and could help improve judge guidelines to better align with human evaluator criteria.

However, we acknowledge that a more extensive study would be required to fully validate LLM judges as reliable examiners or potential copilots for human evaluators.
For example, testing with low-grade responses, which were not available in our collected dataset, could provide more robust comparisons between human and LLM judges. Furthermore, having official examiners to evaluate model-generated responses would improve judge guidelines and achieve a more accurate assessment of LLM performance.

Also, the high cost of using frontier models like o1 remains a significant limitation, although the rapid development of open-source alternatives like DeepSeek-R1 suggests more cost-effective solutions may emerge.

Future work should focus on implementing automated validation mechanisms to ensure scoring accuracy and developing more sophisticated and domain-related prompting techniques that incorporate valuable specialist's feedback. The \emph{oab-bench} and our findings contribute to the broader field of automated evaluation of open-ended tasks in specialized domains.

% This finding is particularly relevant given that human evaluators are prone to errors, as evidenced by the 41\% success rate of grade appeals in recent exams.

% \begin{acks}
% ...
% \end{acks}

\bibliographystyle{ACM-Reference-Format}
\bibliography{references}

\appendix

% \section{Multi-Turn Judge Prompt}
% \label{app:multiturn_prompt}

\section{Disagreement Analysis in Legal Essays}
\label{app:disagreements}

Tables~\ref{tab:scores_criminal_essay} and~\ref{tab:scores_labor_essay} show, respectively, a summarized outline of a grading rubric for Civil Law and Labor Law documents written by humans. In addition to the scores assigned by human evaluators, the table also shows the scores given by LLM judges.

\begin{table}[htb]
\centering
\caption{Analytical Evaluation of Criminal Law. The GPT-4o judge made an error in summing the individual scores, reporting 4.60 when the correct total would be 4.80.}
\label{tab:scores_criminal_essay}
\resizebox{\columnwidth}{!}{
\begin{tabular}{p{7.5cm}ccccc}
\toprule
Item & Value & Human & o1 & GPT-4o & DeepSeek-R1 \\
\midrule
1. Correct addressing: Criminal Special Court of Niterói & 0.10 & 0.10 & 0.10 & 0.10 & 0.10 \\
2. Correct indication of legal provision supporting the criminal complaint & 0.10 & 0.10 & 0.10 & 0.10 & 0.10 \\
3.1. Qualification of complainant and defendant & 0.20 & 0.20 & 0.20 & 0.20 & 0.20 \\
3.2. Existence of Power of Attorney with special powers & 0.30 & 0.30 & 0.30 & 0.30 & 0.30 \\
4.1. Exposition of criminal facts & 0.60 & 0.60 & 0.60 & 0.60 & 0.60 \\
4.2. Description of defamation offense and typical classification & 0.60 & 0.60 & 0.60 & 0.60 & 0.60 \\
4.3. Incidence of penalty enhancement cause & 0.30 & 0.30 & 0.30 & 0.30 & 0.30 \\
4.4. Incidence of formal concurrence of crimes & 0.40 & 0.40 & 0.40 & 0.40 & \textcolor{red}{0.10} \\
5.a. Designation of preliminary or conciliation hearing & 0.20 & 0.20 & \textcolor{red}{0.00} & \textcolor{red}{0.00} & \textcolor{red}{0.00} \\
5.b. Summons of the defendant & 0.20 & 0.20 & 0.20 & 0.20 & 0.20 \\
5.c. Acceptance of the complaint & 0.20 & 0.20 & 0.20 & 0.20 & 0.20 \\
5.d. Hearing of listed witnesses & 0.20 & 0.20 & 0.20 & 0.20 & 0.20 \\
5.e. Conviction of the defendant & 0.90 & 0.90 & 0.90 & 0.90 & \textcolor{red}{0.80} \\
5.f. Setting minimum compensation value & 0.40 & 0.40 & 0.40 & 0.40 & 0.40 \\
6. Witness list: List Carlos, Miguel and Ramirez & 0.20 & 0.20 & 0.20 & 0.20 & 0.20 \\
7. Correct structure & 0.10 & 0.10 & 0.10 & 0.10 & 0.10 \\
\midrule
\textbf{Total} & \textbf{5.00} & \textbf{5.00} & \textbf{4.80} & \underline{\textbf{4.60}} & \textbf{4.40} \\
\bottomrule
\end{tabular}
}
\end{table}

% \begin{table}[htb]
% \centering
% \caption{Analytical Evaluation of Civil Law}
% \label{tab:scores_civil_essay}
% \resizebox{\columnwidth}{!}{
% \begin{tabular}{p{7.5cm}cccc}
% \toprule
% Item & Value & Human & o1 & GPT-4o \\
% \midrule
% 1. Addressing: 15th Civil Court of Rio de Janeiro & 0.10 & 0.10 & 0.10 & \\
% 2. Distribution by dependency & 0.10 & 0.10 & 0.10 & \\
% 3. Name and qualification of parties & 0.20 & 0.20 & 0.20 & \\
% 4. Timeliness & 0.40 & 0.40 & 0.40 & \\
% 5.I. Legitimacy to file third-party motions & 0.60 & 0.60 & 0.60 & \\
% 5.II. Demonstrate that the claimant has marital property rights over the property under execution & 0.80 & 0.00 & \textcolor{red}{0.80} & \\
% 5.III. Characterization of the property as family homestead & 0.60 & 0.50 & \textcolor{red}{0.60} & \\
% 5.IV. Spouse's rights must be protected when property is attached & 0.80 & 0.80 & 0.80 & \\
% 6.a. Proof of court fees payment OR request for free legal aid & 0.10 & 0.10 & 0.10 & \\
% 6.b. Attachment of summary evidence of possession or ownership & 0.30 & 0.00 & \textcolor{red}{0.30} & \\
% 6.c. Production of all applicable means of evidence & 0.10 & 0.10 & 0.10 & \\
% 6.d. Request for granting the motion to declare ineffective/void the attachment regarding marital share & 0.50 & 0.50 & 0.50 & \\
% 6.e. Award of court costs and attorney's fees & 0.20 & 0.20 & 0.20 & \\
% 6.f. Indication of claim value & 0.10 & 0.10 & 0.00 & \\
% 7. Closing & 0.10 & 0.10 & 0.10 & \\
% \midrule
% \textbf{Total} & \textbf{5.00} & \textbf{3.80} & \textbf{4.90} & \\
% \bottomrule
% \end{tabular}
% }
% \end{table}

\begin{table}[htb]
\centering
\caption{Analytical Evaluation of Labor Law. The GPT-4o judge made an error in summing the individual scores, reporting 3.50 when the correct total would be 3.70.}
\label{tab:scores_labor_essay}
\resizebox{\columnwidth}{!}{
\begin{tabular}{p{7.5cm}ccccc}
\toprule
Item & Value & Human & o1 & GPT-4o & DeepSeek-R1 \\
\midrule
1. Complaint addressed to Criminal Court of Cuiabá & 0.10 & 0.10 & 0.10 & 0.10 & 0.10 \\
2. Qualification of parties & 0.20 & 0.20 & 0.20 & 0.20 & 0.20 \\
3. Indication of Art. 847 of CLT & 0.10 & 0.10 & 0.10 & 0.10 & 0.10 \\
4. Preliminary motion of ineptitude for hazard pay claim & 0.50 & 0.50 & 0.50 & 0.50 & 0.50 \\
5. Partial statute of limitations & 0.50 & 0.50 & 0.50 & 0.50 & 0.50 \\
6. Denial of occupational illness compensation & 0.60 & 0.00 & 0.00 & 0.00 & 0.00 \\
7. Dental plan not part of salary & 0.50 & 0.50 & 0.50 & 0.50 & 0.50 \\
8. Food allowance not due for lack of ultra-activity & 0.50 & 0.50 & 0.50 & 0.50 & 0.50 \\
9. Time at disposal (ecumenical service) & 0.50 & 0.50 & 0.50 & 0.50 & 0.50 \\
10. Resignation request / Coercion allegation & 0.50 & 0.00 & \textcolor{red}{0.40} & 0.00 & \textcolor{red}{0.40}\\
11. Functional accumulation & 0.50 & 0.50 & 0.50 & 0.50 & 0.50 \\
12. Renewal of preliminary motion in closing & 0.10 & 0.10 & 0.10 & \textcolor{red}{0.00} & 0.10 \\
13. Renewal of partial statute of limitations & 0.10 & 0.10 & 0.10 & \textcolor{red}{0.00} & 0.10 \\
14. General request for dismissal and evidence indication & 0.20 & 0.20 & 0.20 & 0.20 & 0.20 \\
15. Closing & 0.10 & 0.10 & 0.10 & 0.10 & 0.10 \\
\midrule
\textbf{Total} & \textbf{5.00} & \textbf{3.90} & \textbf{4.30} & \underline{\textbf{3.50}} & \textbf{4.30} \\
\bottomrule
\end{tabular}
}
\end{table}

We emphasize that the official point distribution table, and consequently the official grading rubric, are more detailed. Here we present a simplified version due to space constraints.

In comparing LLM evaluations, we observed that GPT-4o made significant arithmetic errors when summing individual scores in both analyses. Although GPT-4o agreed with o1 on all items of the Criminal law essay, it incorrectly reported a total of 4.60 instead of 4.80. Similarly for the Labor Law essay, it reported 3.50 when the correct sum was 3.70. These basic calculation errors raise concerns about using GPT-4o as a reliable judge for \emph{oab-bench} model responses.

When comparing Human vs. LLM judgments (o1), we found divergence on only one item per essay. DeepSeek-R1 agreed with o1's assessment on these two discrepancies. Upon analyzing these disagreements, the LLM judges' interpretations appear reasonable and well-justified based on the grading criteria. %, suggesting they demonstrate competent evaluation capabilities.

Like LLMs, human evaluators are also susceptible to errors in their assessments. The subjective nature of grading legal essays makes them particularly vulnerable to human factors, such as misinterpreting handwriting, evaluating while fatigued, or confusing elements due to the high volume of exams they must grade. These human limitations can lead to inconsistent or incorrect evaluations, potentially impacting candidates' outcomes.

According to a private law preparatory course that offers a service for grade increase analysis and personalized appeals, 41\% of appeals filed through their service for the 40th bar exam were successful. In other words, 1 in 3 appeals reversed a failing grade~\cite{provadaordem_appeal}. These appeals stem from human grading errors.

% https://www.provadaordem.com.br/blog/post/quais-as-chances-de-um-recurso-da-2a-fase-ser-provido/

\end{document}